%% file: ecai.tex
\documentclass{ecai}
	\usepackage{times}
	\usepackage{graphicx}
	\usepackage{latexsym}

	\usepackage[utf8]{inputenc}
	\usepackage{tikz}
	\usetikzlibrary{arrows}
	\tikzset{
		treenode/.style = {align=center, inner sep=0pt, text centered,
			font=\sffamily},
		arn_n/.style = {treenode, circle, black, font=\sffamily\bfseries, draw=black,
			fill=white, text width=1.5em},
		arn_r/.style = {treenode, circle, red, draw=red, 
			text width=1.5em, very thick},
		arn_x/.style = {treenode, rectangle, draw=black,
			minimum width=0.5em, minimum height=0.5em}
	}
	
	\usetikzlibrary{fit}
	\usetikzlibrary{shapes.geometric}
	\usetikzlibrary{trees}
	\usepackage{verbatim}
	\usepackage{amsthm}
	\newtheorem{definition}{Definition}
	\usepackage[onelanguage, ruled, vlined]{algorithm2e}
	\usepackage{pgfplots}
	\pgfplotsset{compat=newest}
	\newtheorem{example}{Example}

	\newcommand{\as}[1]{\ensuremath{(#1)}}
	\newcommand{\vas}[1]{\ensuremath{[#1]}}
	\newcommand{\wc}[1]{\ensuremath{(#1)}}
	
	\newcommand{\agentview}{{\tt agent view}}

	\newcommand{\StudentAlice}{Student~{\ensuremath{A_1}}}
	\newcommand{\StudentBob}{Student~{\ensuremath{A_2}}}
	\newcommand{\StudentCarol}{Student~{\ensuremath{A_3}}}
	\newcommand{\Alice}{{\ensuremath{A_1}}}
	\newcommand{\Bob}{{\ensuremath{A_2}}}
	\newcommand{\Carol}{{\ensuremath{A_3}}}
	
	\newcommand{\risk}{\ensuremath{{\mbox{\,\it f}}utilityRisk}}

\begin{document}
		
		\title{Utilitarian Distributed Constraint Optimization Problems}
		
		\author{
			Julien Savaux, Julien Vion, Sylvain Piechowiak, Ren\'{e} 
			Mandiau\institute{LAMIH UMR CNRS 8201, University of Valenciennes, France} 
			\and \\
			Toshihiro Matsui\institute{Nagoya Institute of Technology, Japan} 
			\and 
			Katsutoshi Hirayama\institute{Kobe University, Japan} 
			\and 
			Makoto Yokoo \institute{Kyushu University, Japan}
			\and 
			Shakre Elmane, Marius Silaghi \institute{Florida Institute of Technology, USA}
		}
		
		\maketitle
		\bibliographystyle{ecai}
		
		\begin{abstract}			
Privacy has been a major motivation for distributed problem
optimization. However, even though several methods have been proposed
to evaluate it, none of them is widely used. The Distributed
Constraint Optimization Problem (DCOP) is a fundamental model used to
approach various families of distributed problems. As privacy loss
does not occur when a solution is accepted, but when it is proposed,
privacy requirements cannot be interpreted as a criteria of the
objective function of the DCOP.  Here we approach the problem by
letting both the optimized costs found in DCOPs and the privacy
requirements guide the agents' exploration of the search space. We
introduce Utilitarian Distributed Constraint Optimization Problem
(UDCOP) where the costs and the privacy requirements are used as
parameters to a heuristic modifying the search process. Common
stochastic algorithms for decentralized constraint optimization
problems are evaluated here according to how well they preserve
privacy. Further, we propose some extensions where these solvers
modify their search process to take into account their privacy
requirements, succeeding in significantly reducing their privacy loss
without significant degradation of the solution quality.
		\end{abstract}
		
		\section{Introduction}
				
In Distributed Constraint Optimization Problems (DCOP), agents have to
find values to a set of shared variables while optimizing a cost
function.  To find such assignments, agents exchange messages
(frequently assumed to have unspecified privacy implications) to
explore the search space until an optimal solution is found or a
termination condition is met.  Thus, commonly agents reveal
information during the solution search process, causing privacy to be
a major concern in DCOPs~\cite{yokoo1998distributed}.

The artificial intelligence assumption is that utility-based agents
are able to associate each state with a utility
value~\cite{russell2010}. As such, the utility of each action is given
by the difference between the utilities of final and initial states.
If a user is concerned about privacy, then such a user can associate a
utility value with the privacy of each piece of information in the
definition of his local problem. If a user is interested in solving
the problem, he must be also able to quantify the utility he draws
from finding the solution. In a maximization DCOP we assume that the
utility a user obtains from an assignment is represented by the values
of the local constraints of the user for that
assignment. Alternatively, with a minimization DCOP, the constraints
would represent the costs. Certainly, these utilities can be modeled
as a component in a multi-criteria DCOP~\cite{bowring2005distributed}.

Here we approach the problem by assuming that privacy has a utility
that can be aggregated with the utility value for a given DCOP
solution. We evaluate how much privacy is lost by the agents during
the problem solving process, by the total utility of each information
that was revealed.  
For DCOPs with private constraints one assumes that the cost/utility
a constraint associate with a solution, is
the kind of information that the agents would like to keep private.
For DCOPs with privacy of domains, the existence of each value in the
domain of a variable, would be kept private.
For example,
proposing an assignment with that value assigned to the variable has a
privacy cost quantifying the desire of the agent to maintain its
existence
private. While sometimes possibilistic reasoning
was used to guide search~\cite{wallace2004using}, in traditional
algorithms agents explore the search space by proposing values as guided
only by DCOP constraint
costs. 
We propose a new DCOP framework with utility-based agents, where the utility of privacy as well as the utility of each solution is explicitly expressed.
The framework is called Utilitarian Distributed Constraint Optimization Problem (UDCOP).
Simple extensions to
standard stochastic algorithms are studied to verify the impact of this
interpretation of privacy.

Here we evaluate and compare several stochastic algorithms according
to how well they preserve privacy. To do so, we generate distributed
meeting scheduling (DMS) problems, as described
in~\cite{maheswaran2004taking,gershman2008scheduling}. In these
problems, each agent own one variable, corresponding to the meeting to
schedule.
There exists a
global constraint that requires all the variables to be equal, and
also a unary constraint for each agent.

In the next section we discuss existing solvers and approaches to
privacy for DCOPs. Further we formally define the concepts involved in
UDCOPs. In Section~\ref{Algorithms} we introduce some
extensions to common stochastic DCOP solvers that modify the search process to
preserve privacy.  We present our experimental results in
Section~\ref{Experimentations}, before presenting our conclusions.
		
		\section{Background}\label{Background}
		
		Let us first review the most relevant literature concerning DCOPs, stochastic algorithms and privacy measures. 
		
		\subsection{Distributed Constraint Optimization Problems}
		
		Distributed Constraint Optimization Problems (DCOPs)
                have been extensively studied as a fundamental way of
                modeling combinatorial optimization problems in
                multi-agent systems. These problems have been
                addressed with a variety of algorithms, both
                stochastic and systematic.  The systematic techniques
                range from highly asynchronous protocols like
                ADOPT~\cite{modi2005adopt} or asynchronous branch and
                bound~\cite{yeoh2008bnb} to careful constraint
                pseudo-tree traversals like
                DPOP~\cite{petcu2005scalable} or cluster exploitation
                like Asynchronous Partial
                Overlay~\cite{mailler2006asynchronous}.  Algorithms
                like ADOPT are known for their elegant treatment of
                searching within limited bounds from optima, while
                algorithms like DPOP are known for efficiently
                exploiting certain problem structures.  
  The branch and bound
                algorithm~\cite{yeoh2008bnb} keeps expanding nodes in
                the search tree until a solution is found. For
                efficient use of memory, it keeps only the branch from
                the root node to the currently expanded node.
		
 Another common algorithm is Synchronous Branch and Bound (SyncBB)~\cite{hirayama1997distributed}, which was one of the first distributed algorithms for solving DCOPs. SyncBB organizes agents in a chain, and messages can traverse this chain upstream, or downstream. 
Some variations on the basic SyncBB
algorithm include NCBB and AFB~\cite{gershman09,grinshpoun2014privacy}.

DCOP problems have been addressed with constructive
search~\cite{modi2005adopt, brito2009distributed}, fully cryptographic
protocols~\cite{silaghisecure}, or hybrid crypto-constructive
approaches for privacy~\cite{tassa2015max,greenstadt2007ssdpop,grinshpoun2014privacy}.
Researchers have also addressed the issue of objective functions based
on multiple criteria~\cite{clement2013model}, as well as the impact of various aggregation
functions for cost, ranging from the social welfare maximization of
the pure addition to egalitarian leximin~\cite{meisels14,matsui2015leximin}.
		
		

		\subsection{Stochastic Algorithms}

The main stochastic algorithms for solving DCOPs in practice are the
distributed stochastic algorithm, the distributed simulated annealing, D-Gibbs,
and the distributed breakout. In these algorithms,
a flawed solution violating some constraints is revised until all
constraints are satisfied.
		
\paragraph{Distributed Breakout}		
The distributed breakout (DBO)~\cite{yokoo1996distributed} is an
iterative improvement algorithm, originally proposed for DCOPs for
hard constraints (distributed constraint satisfaction problems).  In
DBO, a weight starting at 1 is defined for each pair of assignments
that does not satisfy some constraints.  The evaluation of a given
solution is the summation of the weights of all constraints for the
involved assignment.
With hard constraints, the summation is equal to the number of the constraint
violations.  In the breakout algorithm, an assignment is changed to
decrease the solution value.
		
If the evaluation of the solution cannot be decreased by changing the value of
any variable, the current state may be a local minimum.  When
trapped in a local minimum, the breakout algorithm increases the
weights of constraint violation pairs in the current state by 1 so
that the evaluation of the current state becomes higher than the
neighboring states.  Thus the algorithm can escape from a local
minimum.  Although the breakout algorithm is very simple, it is shown
that it outperforms other iterative improvement algorithms.

\paragraph{Distributed Stochastic Algorithms}
The Distributed Stochastic Algorithm (DSA) is a family of
algorithms~\cite{zhang2002Adistributed}. In DSA, agents start by
randomly selecting an initial value before entering a loop.  In this
loop, each agent first sends its new assigned value (if changed) to
its neighbors, then it collects any new values assigned by those
neighbors. Agents select the next candidate value based on the values
received from other agents, and usually, based also on maximizing some
utility function. The DSA family forms a baseline for evaluating other
algorithms, and there exist a number of
variations~\cite{zhang2002Adistributed} of the DSA algorithm with
slightly different properties. 
These variations differ mainly in the way they choose whether to keep the current state (assignment), or to assign a new one.

Stochastic algorithms are incomplete, namely not guaranteeing
optimality.  Other stochastic algorithms are Distributed Simulated
Annealing and D-Gibbs.  Distributed Simulated
Annealing~\cite{arshad2004distributed} differs from DSA in the way it
picks the next value, and in the use of a schedule of temperatures to
select the probabilities of changes to sub-optimal values.
D-Gibbs~\cite{nguen13} works by mapping DCOPs to probabilistic models
and applies Markov Chain Monte Carlo.


\subsection{Privacy}

Privacy is a fundamental aspect in DCOPs, intrinsic to the main
motivation, in addition to the usual efficiency/optimality trade-offs.
The cost of privacy lost in the process of reaching a solution needs to be
considered~\cite{greenstadt2006analysis}.  For example, in air traffic
control~\cite{international2005worldwide}, each airport has to
allocate take-off and landing slots to the different flights. 
Such coordinated decisions
are in conflict with the need to keep constraints
private~\cite{faltings2008privacy}.

In existing works, several approaches have been developed to deal with
privacy in DCOPs. The first approach using cryptographic techniques
is~\cite{yokoo2002secure}. While ensuring
privacy~\cite{hirt2000efficient}, cryptographic techniques are usually
slower, and sometimes require the use of external servers or computationally
intensive secure function evaluation techniques that may not always be
available or justifiable for their
benefits~\cite{greenstadt2006analysis}. 

Another family of approaches is based on using different search
strategies to minimize privacy loss, as defined by certain privacy
metrics.
		
		\paragraph{Privacy categorization}		
Agents might consider some -or all- of the
following~\cite{grinshpoun2012you} as private information (that they
rather not reveal), and a particular cost could incur in case any of
them is revealed. Types of private information in DCOPs are:
\textbf{domain privacy},
\textbf{constraint privacy},
\textbf{assignment privacy}, and
\textbf{algorithmic privacy}.
		
		
A previously defined framework for modeling privacy requirements with DCOPs
is the Valuations of Possible States (VPS).
VPS~\cite{maheswaran2005valuations,maheswaran2006privacy,greenstadt2006analysis}
measures privacy loss by the extent to which the possible states of
other agents are reduced~\cite{freuder02}.  Privacy is interpreted as
a valuation on the other agents' estimates about the possible states
that one lives in.  During the search process, agents propose their
values in an order of decreasing preference. At the end of the search
process, the difference between the presupposed order of preferences
and the real one observed during search determines the privacy loss:
the greater the difference, the more privacy has been lost.
		
\newcommand{\ssdpop}{
		\paragraph{DPOP with Secret Sharing}
		Distributed Pseudo-tree Optimization Procedure (DPOP)~\cite{petcu2005scalable} 
		is a simple and efficient algorithm with good privacy preservation, 
		along with Adopt (Asynchronous Distributed OPTimization)~\cite{modi2005adopt, silaghi2006nogood, yeoh2008bnb}. 
		The main three phases of DPOP are: 
Creating a Depth-First Search (DFS) tree, where agents sharing constraints 
			are on the same branch. 
Propagating UTIL messages up the tree, starting with
			the leaves.
Determining the optimal values for variables by the root agent.
DPOP with Secret Sharing SSDPOP~\cite{greenstadt2007ssdpop} modifies
DPOP to protect leaves in the DFS tree (that would be subject to
initial vulnerability otherwise) The tree can be viewed as a simple
chain, and an agent $B$ is the bottom of this chain. SSDPOP uses
secret sharing~\cite{shamir1979share} to aggregate the results of a
single meeting $M$, without revealing the individual valuations that
went into that meeting. The aggregate values for meeting $M$ are then
passed to agent $B$. Agent $B$ aggregates this information with his
own valuations and sends the aggregate up the chain. The initial
vulnerability is eliminated.  The SSDPOP algorithm has four phases:
		\begin{itemize}
			\item Creating SSDPOP's topology:  each agent is assigned to a positions in
			the chain. A meeting $M$ is selected for secret sharing.
			\item Secret sharing: The agents in $M$ use secret sharing to send
			aggregate values for $M$ to the bottom agent $B$ without revealing
			their individual valuations. $B$ forms a UTIL message combining its
			valuations with the aggregate valuations for $M$.
			\item UTIL phase: Proceeds just as in DPOP, except that other
			agents in the chain act as if meeting $M$ does not exist.
			\item VALUE phase: Proceeds as in DPOP, except that agent $B$ will
			eventually receive an assignment for all meetings except $M$. When
			$B$ receives the VALUE message from its parent, it computes the
			optimal value for $M$ and sends that value up to the relevant agents.
		\end{itemize}

		\paragraph{Privacy-Preserving Synchronous Branch and Bound}
		Synchronous Branch and Bound (SyncBB)~\cite{hirayama1997distributed}
		was one of the first distributed algorithms for solving DCOPs~\cite{silaghi2006nogood}. 
		Privacy-Preserving Synchronous Branch and Bound 
		(P-SyncBB)~\cite{grinshpoun2014privacy} is a privacy-preserving version
		of SyncBB for solving DCOPs while respecting
		constraint privacy. P-SyncBB preserves the private
		constraint information by computing the costs of CPAs 
		(current partial assignments) and comparing them to the 
		current upper bound, using secure multi-party protocols.
		Some protocols were proposed in~\cite{grinshpoun2014privacy} 
		for this purpose, that can, for example, 
		solve the millionaires' problem securely without resorting to 
		costly oblivious transfer sub-protocols, and compare
		the cost of a CPA, which is shared between two agents, to 
		the upper bound which is held by only one of them.

		\paragraph{Sample Cryptographic Technique}
		In some existing works, cryptographic techniques are used to enforce privacy.
		The main problem
		of these methods is that cryptographic protocols can be much slower,
		which often makes them impractical~\cite{hirt2000efficient}.
		As an example of cryptographic technique, the approach described
		in~\cite{yokoo2002secure}, achieves a high level of privacy using
		encryption, giving more importance to privacy than to the efficiency
		of the resolution. It consists of using randomizable public key
		encryption scheme. In this algorithm, three servers (value selector,
		search controller and decryptor) receive encrypted information from
		agents and cooperate to find an encrypted solution. Relevant parts of
		the solution are then sent to each agent. This method guarantees that
		no information is leaked to other agents. It also guarantees that,
		thanks to the renaming of values by permutation, servers cannot know
		the actual values they are dealing with.
		We now investigate constructive search methods that do not use cryptography.
}

		\section{Concepts}\label{Concepts}
In this section we define formally the distributed constraint optimization
problem, as well as its extensions to utility-based agents. 
		\subsection{Existing Frameworks}
Let us start by presenting the DCOP framework and existing variations.
		\paragraph{Distributed Constraint Optimization Problems}
		The Distributed Constraint Optimization Problem (DCOP) is the
		formalism commonly used to model combinatorial problems distributed
		between several agents.
\begin{definition}
A DCOP is a quadruplet $\langle{A,V,D,C}\rangle$ where:
		
				\begin{itemize}
					\item
					$A=\langle A_1,...,A_n\rangle$ is a vector of $n$ agents
					\item
					$V=\langle x_1,...,x_n \rangle$ is a vector of $n$ variables.  
					Each agent $A_i$ controls 
					the variable $x_i$. 
					\item
					$D=\langle D_1,...,D_n\rangle$ is a vector of domains
					where $D_i$ is the domain for the variable $x_i$, 
					known only to $A_i$, and a subset of $\{1,...,d\}$. 
					\item
					$C=\langle c_1,...,c_m\rangle$ is a vector of weighted constraints, each one defining a cost for each tuple of a relation between variables in~$V$.
				\end{itemize}
				
				The objective is to find an assignment for each variable 
				that minimizes the total cost.
\end{definition}		

\newcommand{\modcop}{
\paragraph{Multi-Objective Distributed Constraint Optimization Problem}
A multi-objective optimization problem (MOOP)~\cite{delle2011bounded} is defined
as the problem of simultaneously maximizing $ k $ objective functions that have no
common measure, defined over a set $ x = \{x_1, . . . , x_N\} $
of $N$ discrete variables, where each $x_j$ takes values in a discrete domain 
$ D_{x_j} = \{ {d^1_j}, ... , {d}^{|D_{x_j}|}_j \} $. 

Thus, a solution to a MOOP is an assignment:

$ a^* = \{ (x_1 = {d}^{(1)}_1 ), ... , (x_N = {d}^{(N)}_N) \} $

of values to variables, such that:
\begin{equation}
	a^*=arg \max\limits_{a \in D_x} U(x) = [U^{-1}(x), ... ,U^k(x)]^T
\end{equation}

where $ D_x ={x^N_{j=1}} D_{x_j} $
is the domain of variables $x$. 

Here, each objective function $ U^i $ can be defined over a subset
$ x_i \subseteq x $ of the variables of the problem. However, for ease
of exposition, we assume each function is defined over the
same set of variables. The choice of the 
metric used to compute $max$ over vectors leads to various types of problems.

A Multi-Objective Distributed Constraint Optimization Problem (MO-DCOP) \cite{clement2013model}
is an extension of the standard mono-objective DCOPs. An MO-DCOP is defined
with a set of agents $S$, a set of variables $X$, multi-objective constraints $C =
\{C^1,...,C^m\}$, i.e., a set of sets of constraint relations, and multi-objective functions
$O = \{O^1,...,O^m\}$, i.e., a set of sets of objective functions. For an objective
$ l (1 \leq l \leq m) $, a cost function $ {f^l}_{i, j} : D_i \times D_j \rightarrow R $, and a value assignment to all
variables $A$, let us denote:

\begin{equation}
	R^l (A) = \sum_{(i,j) \in C^l ,\{(x_i,d_i),(x_j,d_j )\} \subseteq A} 
	{{f^l}_{i, j} (d_i, d_j )}, 
\end{equation}
where $ d_i \in D_i $ and $ d_j \in D_j $
 
Then, the sum of the values of all cost functions for $m$ objectives is defined
by a cost vector, denoted $ R(A)=(R^1(A),...,R^m(A)) $. Finding an assignment
that minimizes all objective functions simultaneously is ideal. However, in general,
since trade-offs exist among objectives, there does not exist such an ideal
assignment. Thus, the optimal solution of an MO-DCOP is characterized by using
the concept of Pareto optimality. Because of this possible trade-off between
objectives, the size of the Pareto front is exponential in the number of agents,
i.e., every possible assignment can be a Pareto solution in the worst case.
}

\begin{example} \label{ex:1}			
Suppose a problem concerning scheduling a meeting between three
students.  They all consider to agree on a place to meet on a given
time, to choose between London, Madrid and Rome. For simplicity, in
the next sections, we will refer to these possible values by their
identifiers: 1, 2 and 3.  The \StudentAlice{} lives in Paris, and it
will cost him $\$70$, $\$230$ and $\$270$ to attend the meeting in
London, Madrid and Rome respectively.  The \StudentBob{} lives in
Berlin, and it will cost him $\$120$, $\$400$ and $\$190$ to attend
the meeting in London, Madrid and Rome respectively.  The
\StudentCarol{} lives in Brussels, and it will cost him $\$40$,
$\$280$ and $\$230$ to attend the meeting in London, Madrid and Rome
respectively.  
The objective is to find the meeting location that minimizes the total cost
students have to pay in order to attend.  

The privacy costs for revealing
her cost for locations 1, 2, and 3 for  \StudentAlice{} are $\$80,\$20,\$40$.
The privacy cost for locations 1, 2 and 3 are $\$100,\$30,\$10$
for \StudentBob{} and $\$80,\$30,\$10$ for \StudentCarol{}.  There exist various reasons
for privacy. For example, students may want to keep their cost for
each location private, since it can be used to infer their initial
location, and they would pay an additional (privacy) price rather than revealing the
said travel cost. 
For example, \StudentAlice{}
associates $\$50$ privacy cost to the revelation of the travel cost of $\$70$ for meeting
in London.
\end{example}

\paragraph{DCOP} The DCOP framework models this problem with:	
\begin{itemize}
			\item
			$A=\{\Alice, \Bob\, \Carol \}$
			\item
			$V=\{x_1,x_2,x_3\}$
			\item
			$D=$ $\{ \{1, 2, 3\}, \{1, 2, 3\} \}$
			\item
			$C= \{ 
\{\as{x_1=1}, 70\}, 
\{\as{x_1=2}, 230\}, 
\{\as{x_1=3}, 270\},$ 
\\
			\hspace*{0.92cm}$ 
\{\as{x_2=1}, 120\}, 
\{\as{x_2=2}, 400\}, 
\{\as{x_2=3}, 190\},$ 
\\ 
			\hspace*{0.92cm}$ 
\{\as{x_3=1}, 40\}, 
\{\as{x_3=2}, 280\},
\{\as{x_3=3}, 230\}
\}$ \\ 
			\hspace*{0.92cm} $\{\neg(x_1=x_2=x_3),\infty\}$
\}
		\end{itemize}
where each constraint is described with the notation $\{p,c\}$ stating that if the predicate $p$ holds then the cost $c$ is payed, and the notation $\as{x=a}$ is a predicate stating that a variable $x$ is assigned a value $a$.

		With a DCOP, costs are paid when a solution is accepted.
		However, privacy costs are already paid whenever the corresponding assignments is proposed.
		This means that privacy costs cannot be interpreted as a criteria of a 
		multi-objective DCOP. With a standard DCOP, agents can explore the search space,
		and then choose the solution with minimal cost. 
		With privacy requirements, the exploration itself is costly, as it implies privacy leaks.
		This means that a given solution may imply different privacy loss depending on 
		the algorithm used to reach the said solution. As it can be observed, 
		DCOPs cannot model the details regarding privacy considerations.

\input{tradeoff.tex}

One could attempt to model the privacy requirements by aggregating the solution quality,
 called $solutionCost$ and the $privacyCosts$ into a unique cost. However, this is not possible:
In a DCOP,  agents explore the search space to find a better solution,
and only pay the corresponding solution cost when the search is over and  the solution 
is accepted. This means that the solution cost decreases with time. However,
privacy costs are cumulative and are paid during the search process itself (each time a solution is proposed), 
no matter what solution is accepted at the end of the computation.
This means that the total privacy loss increases with time (see Figure~\ref{fig:Plot22}).
Aggregating the solution costs and privacy costs or using a multi-criteria DCOP would not 
consider the privacy cost of the solutions that are proposed but not kept as the final one.
		
		\subsection{Extensions}
		\paragraph{Utilitarian Distributed Constraint Optimization Problem}
		We propose to ground the theory of DCOP in the well-principled theory of 
		utility-based agentry.
		%
		We introduce the Utilitarian Distributed Constraint Optimization
		Problem (UDCOP). Unlike previous DCOP frameworks,
		besides results, we are also interested in the search
		process.

		
		\begin{definition}	
			A UDCOP is a tuple $\langle A,V,D,C,U \rangle$ where:
			\begin{itemize}
				\item
				$A=\langle A_1,...,A_n\rangle$ is a vector of $n$ agents
				\item
				$V=\langle x_1,...,x_n \rangle$ is a vector of $n$ variables.  
				Each agent $A_i$ controls 
				the variable $x_i$. 
				\item
				$D=\langle D_1,...,D_n\rangle$ is a vector of domains
				where $D_i$ is the domain for the variable $x_i$, known only to $A_i$, and a subset of 							$\{1,...,d\}$. 
				\item
				$C=\langle c_1,...,c_m\rangle$  is a vector of weighted constraints, each one defining a cost for each tuple of a relation between variables in~$V$.
				\item 
				$U$: a vector of privacy costs for each agent, each one defining the set of costs an agent suffers for the revelation of the values in his variable.
			\end{itemize}
The {\em state} of agent $A_i$ includes the subset of $D_i$ that it has revealed, as well as the cost of the corresponding. 
The problem is to 
search for an assignment of the variables such
such that the total utility is maximized (including privacy and solution utility/cost).

		\end{definition}	
		
		\begin{example}\label{ex:dcop}
			The DCOP in the Example~\ref{ex:1} is extended to a UDCOP 
			by specifying 
			the 
			additional parameter $U$: \\
			$U=\langle\{u_{1,1}=80,u_{1,2}=20, u_{1,3}=40\},\\
			\hspace*{0.7cm} ~~\{u_{2,1}=100,u_{2,2}=30, u_{2,3}=10\}, \\
			\hspace*{0.7cm} ~~\{u_{3,1}=80,u_{3,2}=30, u_{3,3}=10\}\rangle$
		\end{example}
where $u_{i,j}$ is the privacy cost $A_1$ suffers from revealing the assignment $\as{x_i=j}$.

Note that to model this problem with the VPS 
framework, the 3 participants have to suppose an order of
preference between all different possible values for each other
agent. As agents initially do not know anything about others agents
but the variable they share a constraint with, they have to suppose an
equal distribution of all possible values for all other agents, meaning
that they do not expect the feasibility of any value to be less secret, and so proposed
first.  In this direction one needs to extend VPS to be able to also
model the kind of privacy introduced in this example.

\paragraph{UDCOPs with Private Constraints }
If agents are self-interested, each expecting a separate reward from
the solution of the UDCOP, each of them potentially suffering personal
costs described by constraints, and each of them having private costs
for various configuration elements, a further extension would be
needed.
\begin{definition}
A UDCOP with Private Constraints (UDCOPPC) is a tuple $\langle A,V,D,C,U \rangle$ where:
\begin{itemize}
\item
$A=\langle A_1,...,A_n\rangle$ is a vector of $n$ agents
\item
$V=\langle x_1,...,x_n \rangle$ is a vector of $n$ variables.  
\item
$D=\langle D_1,...,D_n\rangle$ is a vector of domains
where $D_i$ is the domain for the variable $x_i$.
\item
$C=\langle C_1,...,C_n\rangle$ is a set of agent constraints, where each
  $C_i=\{c_{i,1},...,c_{i,m_i}\}$ is the set of weighted constraints known to
  agent $A_i$, each one defining a cost or utility for each tuple of a relation between variables in $V$.
\item 
$U=\langle U_1,...,U_n \rangle$: a vector of privacy costs for agents,
each one defining the cost an agent suffers for the revelation of the 
weight it associates with a tuple in some of its constraint.
\end{itemize}
The {\em state} of agent $A_i$ includes the subset of $D_i$ that it
has revealed, as well as the cost of the corresponding. The problem is
to define a set of communication actions
 for each agent such that the total utility is maximized.
\end{definition}

\begin{example}\label{ex:dcoppc}
			
The DCOP in the Example~\ref{ex:1} is extended to a UDCOPPC by modifying the
parameters $C$ and $U$ as follows:
\\
$C_1=
\{ 
c_{1,1}=[\as{x_1=1}, 70], 
c_{1,2}=[\as{x_1=2}, 230],$\\ \hspace*{0.9cm} $
c_{1,3}=[\as{x_1=3}, 270],
c_{1,4}=[\neg(x_1=x_2=x_3),\infty]$\}
\\
$C_2=\{ 
c_{2,1}=[\as{x_2=1}, 120], 
c_{2,2}=[\as{x_2=2}, 400],$\\ \hspace*{0.9cm} $
c_{2,3}=[\as{x_2=3}, 190],
c_{2,4}=[\neg(x_1=x_2=x_3),\infty]$\} 
\\ 
$C_3=\{ 
c_{3,1}=[\as{x_3=1}, 40], 
c_{3,2}=[\as{x_3=2}, 280],$\\ \hspace*{0.9cm} $
c_{3,3}=[\as{x_3=3}, 230],
c_{3,4}=[\neg(x_1=x_2=x_3),\infty]\}
$ \\
%
\\
$U_1=
\{ 
\wc{
c_{1,1},80}, 
\wc{
c_{1,2},20},
\wc{
c_{1,3},40}
$\}
\\
$U_2=\{ 
\wc{
c_{2,1},100}, 
\wc{
c_{2,2},30},
\wc{
c_{2,3},10}$\} 
\\ 
$U_3=\{ 
\wc{
c_{3,1},80}, 
\wc{
c_{3,2},30},
\wc{
c_{3,3},10}
\}
$ 		
\end{example}

A pair $\wc{c,v}$ appearing in the definition of the parameter $U$ specifies that the privacy loss associated with the revelation
of the cost/utility in constraint $c$ is given by $v$.

The fact that a participant expects a reward $r$ for finding a schedule for the
meeting can be modeled in UDCOPPC by replacing the cost of the conflict in the constraint 
$[\neg(x_1=x_2=x_3),\infty]$ from infinity to $r$,
obtaining $[\neg(x_1=x_2=x_3),r]$.
 	
\section{Algorithms}\label{Algorithms}
\LinesNumbered

		Now we discuss how the basic DBO and DSA algorithms are adjusted 
		to UDCOPs. 
		The state of an agent includes the \agentview{}.
		After each state change, each agent computes the estimated 
		utility of the state reached by each possible action, and selects randomly 
		one of the actions leading to the state with the maximum expected utility.
		
		In our algorithms, an information used by agents in their estimation
		of expected utilities is the risk of one of their assignments not being 
		part of the final solution. For each agent $A_i$ can be apriori estimated with the Equation~\ref{eq:rejected}:
		\begin{equation}
		\risk = 1 - \frac{1}{|D_i|}\label{eq:rejected}
		\end{equation}

		Before proposing a new value, agents estimate the utility that will be reached in the 
		next state. This value is the summation of the costs of revealed 
		\agentview{}s (weighted by their probability to be the final solution) in the said state, 
		and of the corresponding privacy costs. 
		
		If this $estimatedCost$ is lower than the estimation of the current state,
		the agent proposes the next value, otherwise it keeps its actual value.

The Distributed Breakout with Utility (DBOU) algorithm is obtained
from DBO by adding the lines 2 to 8 in Algorithm~\ref{alg:DBOU}.
At line~2, the maximal improvement is initialized at 0.
At line~3, the next value is initialized at the current value.
At line~4, the possible next value is set to the value that gives the maximal improvement.
At line~5, the set of revealed values is the union of the already revealed values and the new value
At line~6, we estimate the cost reached after the next value is proposed.
At line~7, the cost of the current state is estimated.
At line~8, if the next cost is lower than the current cost, the maximal improvement and next value are updated

Similarly an algorithm called Distributed Stochastic Algorithm with Utilities (DSAU)
is obtained from DSA, by adding the the lines 6 to 10 in
Algorithm~\ref{alg:DSAU}.
		
		
\begin{example}			
Continuing with Example~\ref{ex:dcop}, at the beginning of the
computation with the DSAU solver, the participants select a random
value. The resulting \agentview{} of each agent is ${x_1=1,x_2=1,x_3=3}$.  The
utilities of the reached state are ${70+u_{1,1}=70+80=150}$,
${120+u_{2,1}=120+100=220}$, and ${230+u_{3,3}=230+10=240}$ for
\StudentAlice{}, \StudentBob{}, and \StudentCarol{} respectively.  The
participants then inform each others of their value. They then
consider changing their value to a new randomly selected one. The
considered \agentview{} is ${x_1=2,x_2=3,x_3=1}$.  If the participants
change their value, the utilities of the reached states would be
${(70+230)/2+u_{1,1}+u_{1,2}=150+80+20=250}$,
${(120+190)/2+u_{2,1}+u_{2,3}=155+100+10=265}$, and
${(40+230)/2+u_{3,3}+u_{3,1}=135+10+80=225}$, for \StudentAlice{},
\StudentBob{}, and \StudentCarol{} respectively.  \StudentAlice{} and
\StudentBob{} do not propose the new value as it would increase their
utility. 

However, \StudentCarol{} chooses to change its value from $2$
to $1$ which lowers its utility from 240 to 225. In the next step, the
\agentview{} is ${x_1=1,x_2=1,x_3=1}$. Participants then do not change
their value anymore, as all other options would not decrease the
utility.  At the final step, the previous \agentview{} is therefore the
optimal solution. With DSAU, the reached utilities are
${70+80=150,120+100=220,40+10+80=130}$ for \StudentAlice{},
\StudentBob{}, and \StudentCarol{} respectively.  With standard DSA,
the final utilities are ${(70+u_{1,1}+u_{1,2}+u_{1,3}=230}$,
${(120+u_{2,1}+u_{2,2}+u_{2,3}=260}$, and
${(40+u_{3,2}+u_{3,1}+u_{3,3}=160}$, for \StudentAlice{},
\StudentBob{}, and \StudentCarol{} respectively.  Therefore, using
DSAU instead of DSA reduces the utility by ${80, 40, 30}$. 
\end{example}
	
\input{DBO.tex}	
\input{DSA.tex}
\input{estimateUtility.tex}

To adapt DBOU and DSAU for privacy of constraints in UDCOPPC, the
revealed domains and possible revealed domains are changed to the
revealed constraints and possible revealed constraints, respectively.


\section{Discussion} 

To further clarify why Multi-Objective DCOPs (MO-DCOPs) cannot integrate our
concept of privacy as one of the criteria they aggregate, we give an
example of what would be achieved with MO-DCOPs, as contrasted with the
results using the proposed UDCOPs.

Note that a MO-DCOP is a DCOP where the weight of each constraint
tuple is a vector of values $\vas{w_i}$, each value $w_i$ representing
a different metric.  Two weights $\vas{w^1_i}$ and $\vas{w^2_i}$ for the
same partial solution, inferred from disjoint sets of weighted
constraints, are combined into a new vector $\vas{w^3_i}$ where each
value is obtained by summing the values in the corresponding position
in the two input vectors, namely $w^3_i=w^1_i+w^2_i$.  The quality of
a solution of the MO-DCOP is a vector integrating the cost of all
weighted constraints. The vectors can be compared using various
criteria, such as leximin, maximin,
social welfare or the Theil index~\cite{meisels14,matsui2015leximin}.

In the following example we show a comparative trace based on one of
the potential techniques in MO-DCOPs, to provide a hint on why
MO-DCOPs cannot aggregate privacy lost during execution in the same
way as UDCOP. In this example, the privacy value of each assignment
and its constraint cost are two elements of an ordered pair defining
the weight of the MO-DCOP. For illustration, in this example pairs of
weights are compared lexicographically with the privacy having
priority.

\begin{table*}
\centering
\begin{tabular}{l|lll|lll}\hline
Framework        & \multicolumn{3}{c|}{UDCOP} & \multicolumn{3}{c}{MO-DCOP}  \\\hline
Agent            & Student $A_1$    & Student $A_2$  & Student $A_3$  & Student $A_1$      & Student $A_2$  & Student $A_3$  \\\hline
value step 1     & 1     & 1   & 3   & 1       & 1   & 3   \\
cost             & 70    & 120 & 230 & 70      & 120 & 230 \\
privacyCost      & 80    & 100 & 10  & 80      & 100 & 10  \\
situation       & 150   & 220 & 240 &
\vas{80,70}\hspace*{-3mm} & \vas{100,120}\hspace*{-4mm} &\vas{10,230}\hspace*{-1mm}
\\\hline
\multicolumn{7}{c}{believed next state}
\\\hline
considered
 & 2     & 3   & 1   & 2       & 3   & 1   \\
cost             & 150   & 155 & 135 & 230     & 190 & 40  \\
privacyCost      & 100   & 110 & 90  & 20      & 10  & 80  \\
situation        & 250   & 265 & 225$^*$ & 
\vas{20,230}$^*$\hspace*{-3mm} & \vas{10,190}$^*$\hspace*{-4mm} &\vas{80,40}\hspace*{-3mm}
\\\hline
\multicolumn{7}{c}{achieved next state}
\\\hline
value step 2     & 1     & 1   & 1   & 2       & 3   & 3   \\
cost             & 70    & 120 & 40  & 230     & 190 & 230 \\
privacyCost      & 80    & 100 & 90  & 100     & 110 & 10  \\
situation        & 150   & 220 & 130 & 
\vas{100,230}\hspace*{-3mm} & \vas{110,190}\hspace*{-3mm} & \vas{10,230}\hspace*{-1mm}
\\\hline
\end{tabular}
\caption{Comparative trace of two rounds with UDCOP DSAU vs. MO-DCOP DSA with lexicographical comparison on vectors, privacy first. Candidate values are marked with $^*$ if they are better than old values, and will be adopted.}\label{tab:modcop}
\end{table*}

\begin{example}			
Suppose we now want to model the Example~\ref{ex:dcop} with a
MO-DCOP. As also illustrated in the trace in Table~\ref{tab:modcop}, at the beginning of the computation with the DSA solver, the
participants select a random value. The resulting \agentview{} is
${x_1=1,x_2=1,x_3=3}$. 
The participants then inform each others of their value. They then consider changing their value to a new randomly selected one. The considered \agentview{} is
${x_1=2,x_2=3,x_3=1}$. 
	
Like with UDCOPs, \StudentAlice{} does not propose the new value as it
would increase their cost, and \StudentCarol{} chooses to change
its variable's value from $2$ to $1$.
	
However, with MO-DCOPs \StudentBob{} changes its value to $3$,
which is not the case with UDCOPs, which implies privacy
loss. The \agentview{} is now ${x_1=1,x_2=3,x_3=1}$.
	
As we see, with the MO-DCOP model, \StudentBob{} reveals more values and 
loses more privacy (with 110-100=10 units of privacy more) than with UDCOPs.
\end{example}

\section{Experimental Results}\label{Experimentations}

We evaluate our framework and algorithms on randomly generated
instances of {\em distributed meeting scheduling problems (DMS)}. 
Previous work~\cite{wallace2005constraint} in distributed
constraint satisfaction problems has already addressed the question of
privacy in distributed meeting scheduling by considering the
information on whether an agent can attend a meeting to be
private. They evaluate the privacy loss brought by an action as the
difference between the cardinalities of the final set and of the initial
set of possible availabilities for a participant. 

The algorithm we use to  generate the problem is:

\begin{enumerate}
	\item We create the variables (one per {\em participant agents}).
	\item We initialize their domain (possible {\em times}).
	\item We add the global constraint {\em ``all equals''}.
	\item Unary constraints are added to variables, to fit the density.
	\item For each unary constraint, we generate a cost between 
	0 and 9. 
	\item For each value of each variable, we generate a
	revelation cost uniformly distributed between 0 and 9.
\end{enumerate}

The experiments are carried out on a computer under Windows 7, using a
1 core 2.16GHz CPU and 4 GByte of RAM. In Figure~\ref{fig:Plot2}, we
show the total amount of privacy lost by all agents, averaged over 50
problems, function of the density of unary constraints. 
In Figure~\ref{fig:PlotTotal}, we show the total cost (solutionCost + privacyCost) for all agents.
Table~\ref{fig:PlotCost} shows that while DBOU is leading to slightly
higher solution cost, the solution cost for DSAU is the same as for DSA.
As we can see, for all problems, the curves of DBOU and DSAU 
are similar to the ones of DBO and DSA, respectively, meaning 
that adding privacy preservation techniques with the UDCOP framework 
does not degrade the quality of the solution.
The problems
are parametrized as follows: 10 agents, 10 possible values, the cost 
for the constraints is a random number between 0 and 9, and the cost
of a revelation is a random number between 0 and 9. Each set of experiments is 
an average estimation over 50 instances with the different algorithms 
(i.e, DBO, DBOU, DSA, DSAU).
	
	\input{plot.tex}
\begin{table}
\begin{center}
\begin{tabular}{|l|c|c|c|c|}\hline
Algorithm & DBO & DBOU & DSA & DSAU\\\hline
Average Solution Quality & 4.33 & 4.55 & 5.11 & 5.11\\\hline
\end{tabular}
\end{center}
\caption{Average solution quality per agent for various algorithms.}\label{fig:PlotCost}
\end{table}
	\input{plotTotal.tex}

\section{Conclusion}\label{Conclusions}
		
While various previous efforts have addressed privacy in distributed
constraint optimization problems, none of the existing techniques is
widely used, likely due to the difficulty in modeling common
problems. As privacy cannot be interpreted as a criteria of a standard
DCOP, we propose in this article a framework called Utilitarian
Distributed Constraint Optimization Problem (UDCOP). It models the
privacy loss for the revelation of an agent's costs for violating
constraints.  We present algorithms that let agents use information
about privacy to modify their behavior and guide their search process,
by proposing values that reduce the amount of privacy loss.  We then
show how adapted stochastic algorithms (DBOU and DSAU) behave and
compare them with standard techniques on different types of
distributed meeting scheduling problems. The experiments show that
explicit modeling and reasoning with the utility of privacy allows
for significant savings in privacy with minimal impact on the quality
of the achieved solutions.
		
		\bibliography{ecai}
		
	\end{document}

%% file: tradeoff.tex
	\begin{figure}[h]
		\centering
		\begin{tikzpicture}
		\begin{axis}[legend cell align=left,
		xlabel={time},
		ylabel={paid cost}
		]
	
		\addplot [color=black, mark= *, solid]coordinates {
			(0.1,8) (0.2,6) (0.3,5)
			(0.4,4) (0.5,3) 
		};
		
		\addplot [color=black, mark= *, dashed]coordinates{
			(0.1,0) (0.2,1) (0.3,2)
			(0.4,4) (0.5,6) 
		};
	
		\addplot [color=black, mark=square, solid]coordinates{
			(0.1,10) (0.2,9) (0.3,9)
			(0.4,10) (0.5,11) 
		};
		
		\pgfplotsset{every axis legend/.append style={
				at={(0.5,-0.2)},
				anchor=north},} 
		
		\legend{privacyCost, solutionCost, totalCost}
		\end{axis}
		\end{tikzpicture}
		\caption{Evaluation of privacy loss on instances with 
			different parameters.}\label{fig:Plot22}
	\end{figure}
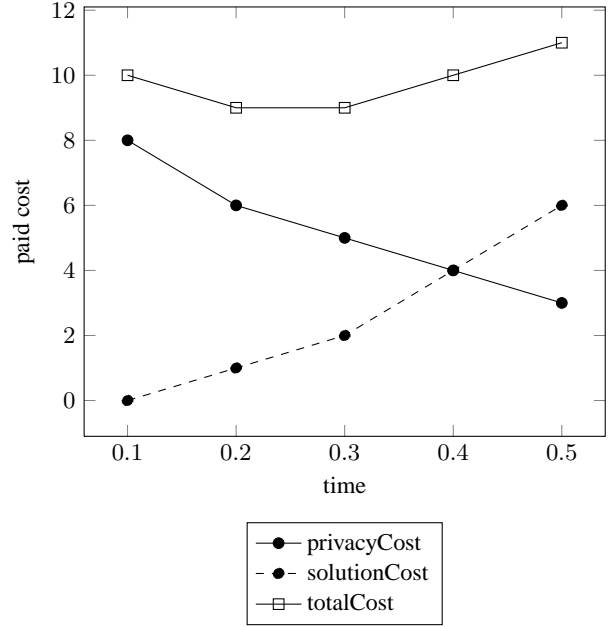

%% file: DBO.tex
\begin{algorithm}[h]

		\KwIn{}
		\KwOut{}
		\SetKwFunction{algo}{algo}\SetKwFunction{proc}{proc}
		\SetKw{KwWhen}{when}
		\SetKw{KwDo}{do}
		currentEval = evaluation value of currentValue\;	
		
		myImprove = 0 \;
		newValue = currentValue \;
		
		possibleValue = the value that gives the maximal improvement\;
		possibleRevealedConstraints = revealedConstraints + constraints containing possibleValue \;
		nextCost =  estimateCost($utilities$, $domain$, $nextRevealedValues$) \;
		currentCost = estimateCost($utilities$, $domain$, $revealedValues$) \;
		\If {(nextCost $<$ currentCost)}
				{
					myImprove = possible max improvement \;
					newValue = the value that gives the maximal improvement\;
				}
		\If {currentEval = 0}
		{consistent = true}
		\Else{
			consistent = false \;
			myTerminationCounter = 0\;}
		\If {myImprove $>$ 0}
	{canMove = true\;
	quasiLocalMinimum = false\;}
	\Else	
	{canMove = false\;
	quasiLocalMinimum = true\;}
	send (improve, $x_i$, myImprove, currentEval, myTerminationCounter) to neighbors \;
	\setcounter{AlgoLine}{0}
	\caption{Procedure sendImprove in DBOU}\label{alg:DBOU}

\end{algorithm} 

%% file: DSA.tex
\begin{algorithm}[h]

		\KwIn{}
		\KwOut{}
		\SetKwFunction{algo}{algo}\SetKwFunction{proc}{proc}
		\SetKw{KwWhen}{when}
		\SetKw{KwDo}{do}
		Randomly choose a value\;
		\While{no termination condition is met} 
		{
		\If{a new value is assigned}
		{send the value to neighbors \;}	
		collect neighbors' new values, if any\;
		
		possibleValue = randomly choose a value \;
		possibleRevealedConstraints = revealedConstraints + constraints containing possibleValue \;
	nextCost =  estimateCost($utilities$, $domain$, $nextRevealedValues$) \;
	currentCost = estimateCost($utilities$, $domain$, $revealedValues$) \;
	\If {(nextCost $<$ currentCost)}
		{
		assign possibleValue;
		}
		}
		\setcounter{AlgoLine}{0}
	\caption{DSAU algorithm}\label{alg:DSAU}

\end{algorithm} 

%% file: estimateUtility.tex
\newcounter{algoline}
\newcommand\Numberline{\refstepcounter{algoline}\nlset{\thealgoline}}
\begin{algorithm}[h]
	\KwIn{$utilities$, $domain$, $revealedValues$}
	\KwOut{$estimatedCost$}
	cost = 0 \;
	privacyCost = 0 \;
	\ForEach{value v in $domain$}
	{\ForEach{constraint c in $constraints$}
		{
			\If{((c contains the assignment of v to $x_i$) and (v is in $revealedValues$))}	
			{
				cost += (utilities.getCost(c) / (domain size of $x_i$) \;
				privacyCost += privacyCost of c \;
			}

		}
	}
	estimatedCost = cost + privacyCost \;
	return estimatedCost \;
	\caption{estimateCost}\label{alg:estimateUtility}  
\end{algorithm}

%% file: plot.tex
	\begin{figure}[h]
		\centering
		\begin{tikzpicture}
		\begin{axis}[legend cell align=left,
		xlabel={Problem Density},
		ylabel={Privacy Loss per Agent}
		]
	
		\addplot [color=black, mark= *, solid]coordinates {
			(0.1,0.05) (0.2,0.19) (0.3,0.14)
			(0.4,0.39) (0.5,0.48) 
		};
		
		\addplot [color=black, mark= *, dashed]coordinates{
			(0.1,0.13) (0.2,0.30) (0.3,0.53)
			(0.4,0.51) (0.5,0.84) 
		};
	
		\addplot [color=black, mark=square, solid]coordinates{
			(0.1,0.73) (0.2,1.75) (0.3,2.46)
			(0.4,3.22) (0.5,4.22) 
		};
		
		\addplot [color=black, mark=square, dashed]coordinates{
			(0.1,0.73) (0.2,0.99) (0.3,1.10)
			(0.4,1.28) (0.5,1.47)
		};

		\pgfplotsset{every axis legend/.append style={
				at={(0.5,-0.2)},
				anchor=north},} 
		
		\legend{DBO, DBOU, DSA, DSAU}
		\end{axis}
		\end{tikzpicture}
		\caption{Evaluation of privacy loss on instances with 
			different densities and algorithms.}\label{fig:Plot2}
	\end{figure}
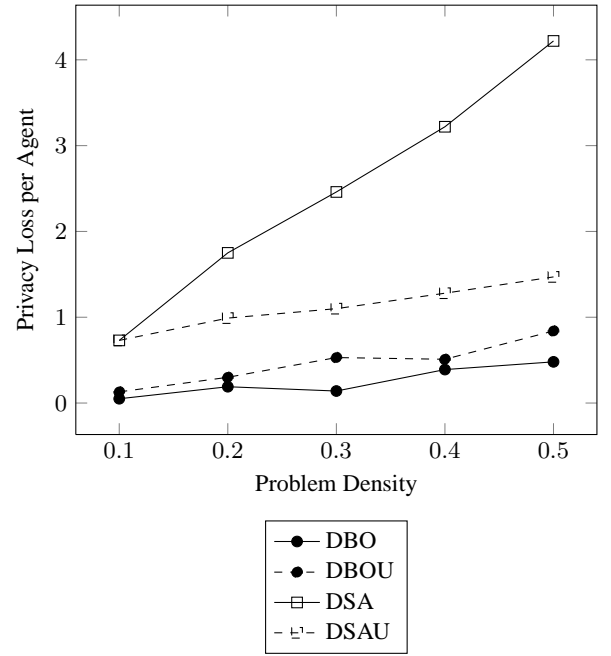	

%% file: plotTotal.tex
	\begin{figure}[h]
		\centering
		\begin{tikzpicture}
		\begin{axis}[legend cell align=left,
		xlabel={Problem Density},
		ylabel={Total Cost per Agent}
		]
	
			\addplot [color=black, mark= *, solid]coordinates {
				(0.1,4.38) (0.2,4.52) (0.3,4.47)
				(0.4,4.72) (0.5,4.81) 
			};
			
			\addplot [color=black, mark= *, dashed]coordinates{
				(0.1,4.68) (0.2,4.85) (0.3,5.08)
				(0.4,5.06) (0.5,5.39) 
			};
			
			\addplot [color=black, mark=square, solid]coordinates{
				(0.1,5.84) (0.2,6.86) (0.3,7.57)
				(0.4,8.33) (0.5,9.33) 
			};
			
			\addplot [color=black, mark=square, dashed]coordinates{
				(0.1,5.84) (0.2,6.10) (0.3,6.21)
				(0.4,6.39) (0.5,6.58)
			};

		\pgfplotsset{every axis legend/.append style={
				at={(0.5,-0.2)},
				anchor=north},} 
		
		\legend{DBO, DBOU, DSA, DSAU}
		\end{axis}
		\end{tikzpicture}
		\caption{Evaluation of total cost on instances with 
			different densities and algorithms.}\label{fig:PlotTotal}
	\end{figure}
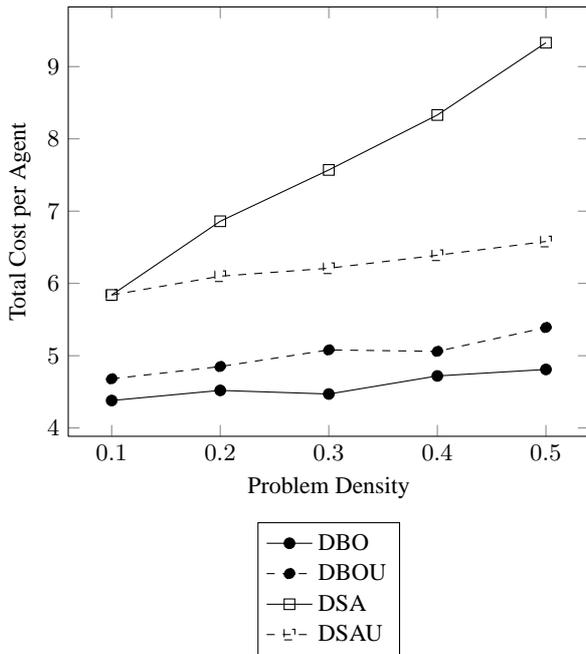